\documentclass{article}

 \usepackage[main, final]{neurips_2026}

\usepackage[utf8]{inputenc} 
\usepackage[T1]{fontenc}    
\usepackage{hyperref}       
\usepackage{url}            
\usepackage{booktabs}       
\usepackage{amsfonts}       
\usepackage{nicefrac}       
\usepackage{microtype}      
\usepackage{xcolor}         
\usepackage[noend]{algpseudocode}
\usepackage{algorithm}
\usepackage[noend]{algpseudocode}
\usepackage{algorithm}
\usepackage{microtype}
\usepackage{graphicx}
\usepackage{subcaption}
\usepackage{booktabs} 
\usepackage{amsmath}
\usepackage{amssymb}
\usepackage{mathtools}
\usepackage{amsthm}
\usepackage{url}
\usepackage{multirow}
\usepackage{multicol}
\usepackage{float}      
\usepackage{wrapfig}
\usepackage{wrapstuff}

\usepackage{hyperref}

\usepackage{amsmath,amsfonts,bm}

\def\eqref#1{equation~\ref{#1}}

\def\1{\bm{1}}

\def\vh{{\bm{h}}}

\def\vx{{\bm{x}}}

\def\mW{{\bm{W}}}

\DeclareMathAlphabet{\mathsfit}{\encodingdefault}{\sfdefault}{m}{sl}
\SetMathAlphabet{\mathsfit}{bold}{\encodingdefault}{\sfdefault}{bx}{n}

\def\gV{{\mathcal{V}}}

\newcommand{\KL}{D_{\mathrm{KL}}}

\theoremstyle{plain}

\theoremstyle{definition}

\theoremstyle{remark}

\newcommand\method{\text{ADJUST}}
\newcommand\mask{\textup{M}}

\title{Enabling Approximate Joint Sampling in Diffusion LMs}

\author{%
  Parikshit Bansal \thanks{parikshitb52@gmail.com}\\
  UT Austin\\
  \And
  Sujay Sanghavi \\
  UT Austin \\
}

\begin{document}

\maketitle

\begin{abstract}
    In autoregressive language models, each token is sampled by conditioning on all the past tokens; the overall string has thus been sampled from the correct underlying joint distribution represented by the model.
    In contrast, masked diffusion language models generate text by unmasking tokens out of order and potentially in parallel. Generating an overall string sampled from the correct underlying joint distribution would (again) require exactly one token unmasking in every full-model forward pass. The more tokens unmasked in parallel, the further away the string is from the true joint; this can be seen in the resulting drop in accuracy (but, increase in speed).
    In this paper we devise a way to {\em approximately} sample multiple tokens from the joint distribution in a single full-model forward pass; we do so by developing a new lightweight single-layer ``sampler" on top of an existing large diffusion LM. One forward pass of the full model can now be followed by multiple forward passes of only this sampler layer, to yield multiple unmasked tokens. Our sampler is trained to mimic exact joint sampling from the (frozen) full model.
    We show the effectiveness of our approximate joint sampling for both pretrained-only (Dream-7B-Base, Llada-7B-Base) and instruction-tuned (Dream-7B-Instruct, Dream-7B-Coder) models on downstream tasks (GSM8k, MATH, MBPP, HEval) and on unconditional language modeling. When eight tokens are unmasked for each full-model denoising step, our sampling algorithm achieves a MAUVE score of 0.84 (vs marginal baseline of 0.19) with respect to the true joint distribution.
\end{abstract}

\section{Introduction}
Masked diffusion language models~\cite{sahoo2024simple,austin2021structured, lou2023discrete} involve generating text strings by starting from an all-masked sequence of tokens, and then iteratively replacing the masked tokens with tokens from the vocabulary, with each ``denoising" forward pass unmasking one or a few tokens. As opposed to auto-regressive models which generate tokens left to right and one token in each forward pass, in masked diffusion models tokens can be potentially unmasked in any order and also potentially multiple tokens can be unmasked in parallel. 

The higher the number of tokens unmasked in parallel after a single denoising forward pass, the faster and cheaper the overall generation~\cite{sahoo2024simple}. However, the quality of the resulting generation is also generally seen to be worse. State-of-the-art large masked diffusion models (7B parameters) are advertised~\cite{ye2025dream,nie2025large,song2025seed,khanna2025mercury} to achieve competitive downstream task performance, when compared to their auto-regressive counterparts; however, this parity is only achieved via slow one-token-per-denoise-step generation from the masked diffusion model. On decreasing the number of denoising steps, there is a sharp decrease in their performance. For example, Dream-7B-Instruct model's accuracy drops from 85\% accuracy to 77\% when generating two tokens instead of one per denoising step~\cite{israel2025accelerating}.

\begin{figure*}[t]
    \centering
    \begin{subfigure}{0.49\linewidth}
        \centering
        \includegraphics[width=\linewidth]{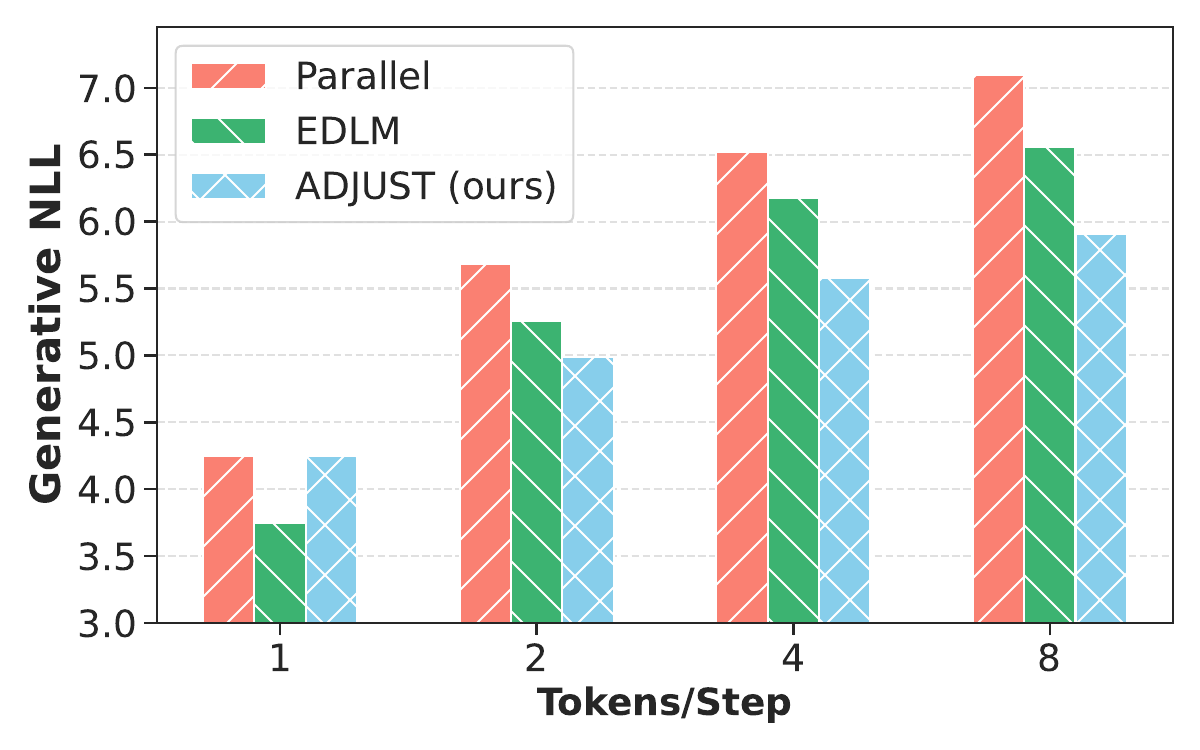}
    \end{subfigure}
    \hfill
    \begin{subfigure}{0.49\linewidth}
        \centering
        \includegraphics[width=\linewidth]{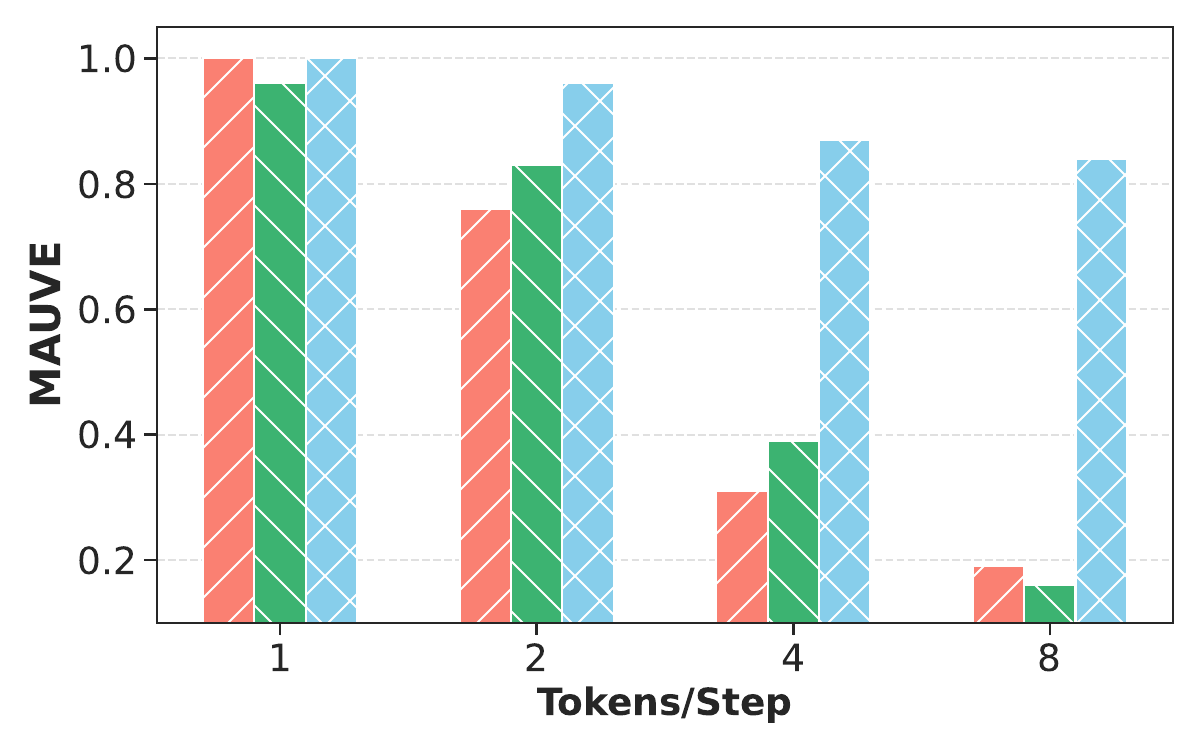}
    \end{subfigure}
    \caption{\label{fig:hero} In this figure, we use Dream-7B-Base to generate a token string of length 128 tokens starting from an all masked string (kindly refer to Sec~\ref{subsec:language} for details). We report the generative negative log-likelihood (lower is better) and the MAUVE score (higher is better) for the generated strings. We vary the number of tokens sampled per diffusion denoising step from one to four. 
    We observe an increase in gen. NLL and a decrease in the MAUVE score as more tokens are generated in parallel. 
    Our method \method\ achieves the best NLL and MAUVE (as compared to baseline methods like naive parallel sampling, and energy-based models). 
    }
\end{figure*}

Our goal is to alleviate the drop in performance seen during parallel generation, while still unmasking multiple tokens in each denoising forward pass of the full masked diffusion language model. Our {\bf main intuition} is that a big reason vanilla parallel generation is inaccurate because it is sampling multiple tokens from the product of their conditional marginal distributions, instead of sampling them from the true joint distribution the model encodes. In particular, the output token probabilities from the forward pass of diffusion model represent per-token marginals, and vanilla parallel generation involves sampling multiple of these independent of each other (we formalize this intuition later). This stands in contrast to auto-regressive models, where every token is correctly conditioned on all the ones generated before it -- left-to-right generation in auto-regressive models samples from the true distribution. 

Similar exact sampling in masked diffusion language models would require exactly one token unmasked in every forward pass. Our {\bf main contribution} is devising a new way to do {\bf approximate sampling} of multiple tokens from their joint distribution -- by unmasking them one at a time using a small auxiliary ``approximate sampler'' model that undertakes many forward passes after every single denoising forward pass of the main big masked diffusion language model. This allows the sampling of our tokens to be at least informed by the specific identities of the other tokens unmasked before them in the same big-model forward pass.


We term our method as \method\ (Sec~\ref{subsec:joint_sampler}). 
\method\ is a {\bf baby version of the base diffusion model}, that is added on to the frozen base model. Our model reuses the hidden states of the base model, and is trained to so that its forward passes mimic those of the base model.
Importantly, \method\ is not a separate distilled diffusion model; rather, it is designed to work together with the base model. 
Algorithmically, each diffusion denoising step with access to \method\ corresponds to a single forward pass of the base diffusion, followed by $K-1$ forward passes of the \method\ module. 
To work in tandem with the base diffusion model, \method's architecture and training recipe have been carefully designed. The objective of all of these is for the output of \method \, to mimic true (slow) joint sampling as closely as possible.



Contributions: 
\begin{enumerate}
    \item Each denoising forward pass of a masked diffusion model provides every token a marginal distribution for that token given the current input; there is no way to jointly sample even two tokens with one forward pass (Sec~\ref{subsec:problem}). True joint sampling is possible only if exactly one token is unmasked in every denoising step. Evidence of the value of joint sampling can be seen in existing work, where it shows up as increased accuracy with increased denoising steps. We develop a new architecture that enables approximate joint sampling of a few tokens at a time for an existing off-the-shelf masked diffusion model. Section~\ref{subsec:joint_sampler} develops the probabilistic view motivating our method.
    \item Our approximate joint sampler consists of a single layer ``on top of'' the diffusion model; we call this \method. Approximate joint sampling is achieved by executing multiple forward passes of only this small \method, for a single forward pass of the big base model. Each forward pass of the \method\ augments the input to the next forward pass of the \method. Section~\ref{subsec:joint_sampler} describes in detail our architecture. 
    \item Training the single-layer sampler model involves a series of non-trivial choices guided by our probabilistic view. This is because each forward pass of the small sampler needs to mimic an actual forward pass of the full model. This needs realistic inputs to the drafter at various noise levels, which in turn have to be offline generated by the base model. Section~\ref{subsec:training} provides the details of how this is achieved.
    \item We show the utility of our method \method\ on both \textit{unconditional generation} (Sec~\ref{subsec:language}) and downstream evaluation (Sec~\ref{subsec:downstream}). Across models, sampling parameters and evaluation settings, \method\ yields consistent gains compared to naive parallel sampling. For example, when considering four tokens generated per time step, for the Dream-Base model, \method\ increases GSM8k accuracy by 20 percentage points and MAUVE score by 0.5 points. In the absence of domain-specific prompts, a generically trained (assuming input prompt to be null) \method, is able to generalize and perform well on downstream tasks as well.
    
\end{enumerate}

\section{Related Work}

Discrete diffusion language models (LMs)~\cite{sahoo2024simple,shi2024simplified,lou2023discrete} are a promising alternative to auto-regressive (AR) LMs. Large diffusion LMs~\cite{ye2025dream,nie2025large} competing with the AR counterparts are based on an absorbing-state Markov process, which noises tokens into a ``masked" state. 
~\cite{zheng2024masked} show that masked diffusion language models can also be considered as mask denoisers.
~\cite{rout2025anchored} show that latent conditioning on important "anchor" tokens leads to much better performance.  

In order to further the capabilities of diffusion LMs~\cite{liu2024discrete, israel2025accelerating} propose to use both AR and diffusion models in tandem to get final logits. 
~\cite{israel2025accelerating, guo2025reviving} propose training-free decoding methods for any-order models with the goal of improving token latency. 
Other works ~\cite{wu2025fast} introduced KV cache for masked diffusion LMs and adaptively decide the number of tokens to decode per denoising step. 
~\cite{campbell2025self, arriola2025encoder} train specialized models which are cheaper to infer. 
Our work instead proposes parallel decoding as joint sampling from the underlying distribution of diffusion model and proposed to train a baby model to work in tandem for the task. 

\section{Method}\label{sec:method}

{\bf Notation}
We denote our input $\vx = [x^{0},x^{1}, \ldots, x^{L-1}] \in \gV^L$ as a string of $L$ tokens in vocabulary $\gV$. We let $\mask \in \gV$ denote a special mask token in the vocabulary. Positions having \mask\ token are considered ``missing" and have to be appropriately filled in with non-mask tokens in $\gV/\mask$. We let $\mathcal{M}(\vx)$ denote the indices (in string $\vx$) of these masked positions. 
Let $\vx \oplus x^i$ denote a new string where the $i$th position of $\vx$ (implicitly assumed to be a masked position) is filled with the token value $x^i$. 
Let masked diffusion models be represented by the pair $(f,\mW)$.
Here $f:\gV^L\to\mathbb{R}^{d\times L}$ \footnote{we assume that the time information is implicit in the masked input sequence~\cite{zheng2024masked}} maps from an input string of $L$ tokens $\vx$ to an output of $L$ embeddings, with $f^i(\vx)$ denoting the output embedding at the $i^{th}$ position. $\mW \in \mathbb{R}^{\gV \times d}$ denotes the language model (LM) head that maps an embedding to logits. Additionally, let $p^i(\cdot\mid\vx)$ denote the marginal probabilities at position $i$ conditioned on $\vx$ and $p_*$ denote the true underlying distribution of the diffusion model (we develop these concepts in Sec~\ref{sec:method}). 

\textbf{Background.}
We study masked diffusion models with ``unmask and commit'' inference; that is, the tokens to be generated are initially started out as $\mask$, and these locations are iteratively filled in with other tokens from the vocabulary via forward passes of the model. Unconditional generation would thus involve a series of strings $\phi \rightarrow \vx_1 \rightarrow \vx_2 \rightarrow \ldots \rightarrow \vx_T$, where $\phi = \vx_0$ is the all-mask sequence and $\vx_T$ is the completely unmasked string. 



\begin{figure*}[t]
    \centering
    \includegraphics[width=0.95\linewidth]{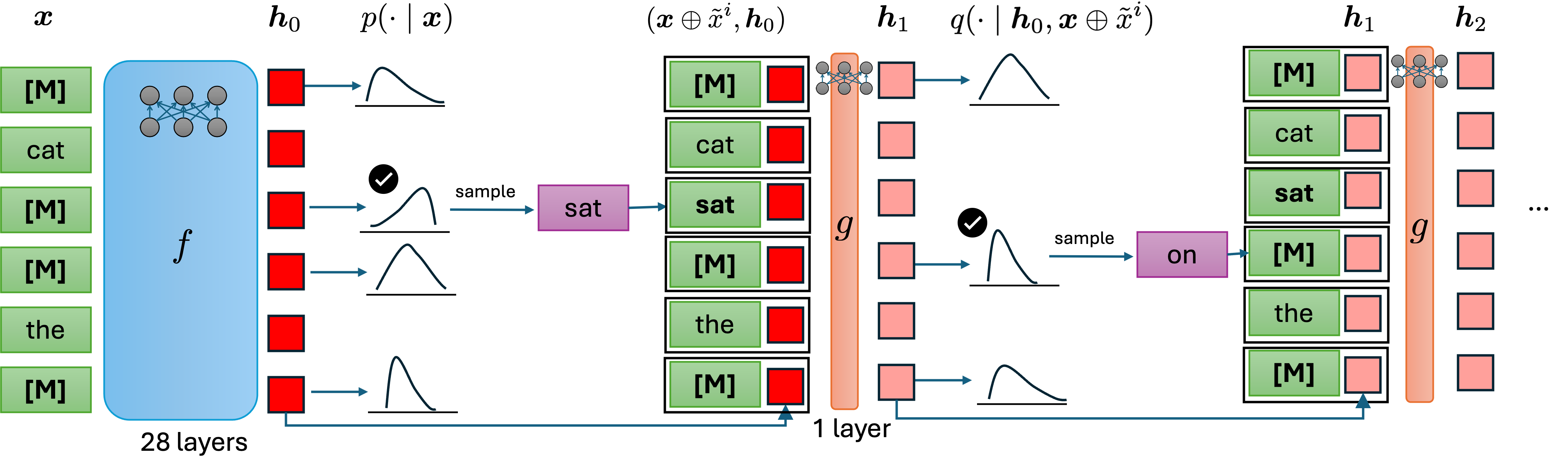}
    \caption{
    \textbf{ADJUST.}
    The figure shows one forward pass of the base diffusion models followed by two passes of the \method\ sampling module.
    The tick in the figure denotes the position selection logic (TOP in Algo~\ref{alg:joint_sampling}). \method\ module $g$ is a single layer bi-direction attention based diffusion LM. After unmasking a single token (\texttt{sat} in the figure), the module $g$ conditions the prior output embeddings $\vh_0$ on the sampled token before proceeding to sample the next token \texttt{on}. Notation in figure follows Algo~\ref{alg:joint_sampling}.
    }
    \label{fig:sampler}
\end{figure*}

\subsection{Problem Setup}\label{subsec:problem}
Let $(f,\mW)$ denote a frozen base masked diffusion model. We will make a joint sampler for $(f,\mW)$.
Recall that discrete diffusion models only output per-token marginals~\cite{sahoo2024simple, austin2021structured}; that is for each position $i = 1,\ldots,L$ a single forward pass yields a probability distribution  $p^i(\cdot|\vx) = \mathrm{softmax}[\mW f^i(\vx)]$ over the vocabulary $\mathcal{V}$. If multiple tokens are unmasked in the same forward pass, this is done by independently generating each of them from their respective marginals. Furthermore, let $p_*$ denote the model's true underlying joint distribution. Intuitively, for a perfectly trained model this is data distribution used to train $(f,\mW)$.  

\textbf{True underlying joint distribution}. 
If we consider the \mask\ tokens as nulls, then for a fixed output position $i$ we can interpret the model's output $p^i(x^i|\vx)$ as the conditional distribution of that token $i$, given the unmasked tokens in $\vx$; thus, for any single position $i$, $p_*(x^i|\vx) = p^i(x^i|\vx)$. 
Then, exact unconditional generation of a length-$L$ fully unmasked string $\vx$ from this $p_*$ can be written as the following via chain rule (with $\sigma$ as some permutation of $[L]$):
\begin{eqnarray}\label{eqn:true_joint}
    p_*(\vx) = \prod_{k=1}^L \,  p_*(x^{\sigma(k)}\, | \, \phi \oplus  \ldots \oplus x^{\sigma(k-1)}) = \prod_{k=1}^L \, p^{\sigma(k)}(x^{\sigma(k)}\, | \, \phi\oplus  \ldots \oplus x^{\sigma(k-1)})  
\end{eqnarray}
However, this corresponds to sampling {\em exactly} one token at a time -- with the unmasking order given by  $\sigma$ of the $L$ positions. 

{\bf The product marginal problem.} While one-token-per-forward-pass unmasking represents exact sampling, it is also very slow and expensive; existing work advocates sampling multiple tokens in parallel with one forward pass. However, this corresponds to sampling from the product of their individual marginals; for example, if $K$ tokens $x^{\sigma(1)},\ldots,x^{\sigma(K)}$ are generated in parallel in the first step after $\phi$, their joint distribution satisfies
\begin{eqnarray}\label{eqn:parallel}
    p_{parallel} \left ( x^{\sigma(1)},\ldots,x^{\sigma(K)} \right ) & = & \prod_{k=1}^K \, p^{\sigma(k)} \left ( x^{\sigma(k)} \, | \, \phi  \right )
\end{eqnarray}
Thus, parallel generation results in a sequence drawn from a different distribution from the true $p_*$; our paper is based on the idea that this is the cause of the (widely recognized) loss in accuracy when multiple tokens are sampled in parallel.

\subsection{Our method : \method}\label{subsec:joint_sampler}


We develop a new method to generate $K$ tokens with one forward pass of the base diffusion model $(f,\mW)$. We do this by sampling one token at a time from a {small extra ``baby diffusion'' draft model (see Fig~\ref{fig:sampler}). Crucially, mimicking one token at a time sampling, the token unmasked after each step of \method \, is fed back in to the input for the next step of \method. 
This allows for each unmasking to be informed by the identities of the other tokens unmasked before it; in contrast, simple parallel sampling results in each token being sampled independent of the identities of the other tokens. We illustrate this in Fig~\ref{fig:sampler}. Importantly, \method\, conditions on the tuple of updated token string and most recent output embeddings (elaborated on later). This enables ADJUST to work in tandem with the base model and differentiates it from a stand-alone distilled diffusion LM.  

\paragraph{\method\ } The \method\ model is represented as the pair $(g,\mW)$. Different from $f$, the function $g$ operates on a pair of inputs $g : \mathbb{R}^{d\times L} \times \gV^L \to \mathbb{R}^{d\times L}$. Let $\vh \in \mathbb{R}^{d\times L}$ denote the set of ``most current'' embeddings (Sec~\ref{subsec:inference}) and let $\vx$ be the current unmasked string. 
Then for each position $i$, \method\ yields a probability distribution :
\begin{align*}
    q^i \left ( \cdot \mid \vh, {\vx} \right ) = \text{softmax}\big[\mW g^i(\vh, {\vx}) \big]
\end{align*}
We draw the reader's attention to the parallels between $q$ and $p$, where the only difference between the two is the inclusion of the extra embeddings $\vh$ as input for $g$. This allows $g$ to be much smaller model than $f$.

\begin{figure}[t]
\centering
\begin{minipage}{0.48\textwidth}
\begin{algorithm}[H]
\caption{\label{alg:dafter_training} \method\ Training (Sec~\ref{subsec:training})}
\begin{algorithmic}[1]
\State {\bf Input:}  $\vx \sim \mathcal{D}$ input strings, 
(frozen) diffusion model $f,\mW$ (where $\mW$ is the LM head)
\Repeat
    \Statex \textsc{\textcolor{brown}{Sample training strings (Sec~\ref{subsec:training})}}
    \State $\vx \leftarrow$ final string after a random number of denoising steps by base model on $\vx$
    \State $\vx_{1},\ldots,\vx_{K} \leftarrow$ strings after further $K$ denoising steps by base model on $\vx$
    \State $\vh_{1},\ldots, \vh_{K} \gets f(\vx_{1}),\ldots, f(\vx_{K})$ 
    \State $\hat{\vh}_0 \gets f(\vx)$
    \Statex \textsc{\textcolor{brown}{Unroll embeddings (Eqn~\ref{eqn:unroll})}}
    \For{$k = 0$ to $K-1$}
        \State $\hat{\vh}_{k+1} \gets g_\theta(\hat{\vh}_{k},\vx_{k+1})$ \label{algo_eqn_unrolling}
    \EndFor
    \Statex \textsc{\textcolor{brown}{Compute loss (Eqn~\ref{eqn:training_loss})}}
    \State $\mathcal{L}(\theta) \gets \sum^{K}_{k=1} \KL(\mW\vh_{k} \mid\mid \mW\hat{\vh}_{k})$ \label{algo_eqn_loss}
    \State Take gradient step on $\nabla_\theta \mathcal{L}(\theta)$
\Until{converged}
\State {\bf Return:} $g_\theta$
\end{algorithmic} 
\end{algorithm}
\end{minipage}
\hfill
\begin{minipage}{0.48\textwidth}
\begin{algorithm}[H]
\caption{\label{alg:joint_sampling} \method: Sampling (Sec~\ref{subsec:inference})}
\begin{algorithmic}[1]
\State \textbf{Input:} Prompt $\vx$, frozen discrete diffusion model $f,\mW$ (where $\mW$ is the LM head), \method\ $g$, number $K$ of tokens to generate in each step, base model's unmasking logic TOP
\While{$|\mathcal{M}(\vx)| > 0$}
    \State $\vh_{0} \gets f(\vx)$ 
    \For{$k = 0$ to $K-1$}
        \Statex \textsc{\textcolor{brown}{Select position to unmask}}
        \State $i \gets {\mathrm{TOP}}\{\mW{\vh}^j_{k} : j\in \mathcal{M}(\vx)\}$ 
        \Statex \textsc{\textcolor{brown}{Sample token, update token string}}
        \State $\tilde{x}^{i} \sim \mathrm{softmax}(\mW{\vh}^i_{k})$ 
        \Statex \textsc{\textcolor{brown}{Get update token string}}
        \State $\vx \gets \vx \oplus \tilde{x}^{i}$ 
        \Statex \textsc{\textcolor{brown}{Update $\vh$ on token $\tilde{x}^i$ (Eqn~\ref{eqn:unroll})}}
        \State \textcolor{orange}{${\vh}_{k+1} \gets g({\vh}_{k},\vx)$} \label{diffline}
    \EndFor
\EndWhile
\State \textbf{Return:} $\vx$ 
\end{algorithmic}
\end{algorithm}
\end{minipage}
\end{figure}

\textbf{Architecturally}, \method\ first embeds the input token string $\tilde{x}^{\sigma(1)} \oplus \ldots \oplus \tilde{x}^{\sigma(k-1)}$ using the frozen base model's token embeddings. It then concatenates these token embeddings with the set of embeddings $\vh$ given as input. The concatenated embeddings are then down-projected into the base model's hidden dimension size and then passed through a transformer decoder layer. For simplicity, \method\ model's decoder layer has the same architecture as the base model's decoder layers. We refer the reader to Fig~\ref{fig:joint_architecture} (in appendix) for an architecture diagram of $g$.

\subsection{Inference with \method} \label{subsec:inference}
Let $\vx$ be the current unmasked string and suppose we are sampling $K$ tokens per denoising step. Let $\sigma(1),\ldots,\sigma(K)$ be the set of positions of the tokens that will be sampled. We now describe how inference proceeds with ADJUST. 

\textbf{Current Embeddings} are the output embeddings of the most recent forward pass of either the base model $f$ or ADJUST module $g$. Concretely, let $\vh_0 = f(\vx)$, define the most recent embedding after doing a forward pass with the base diffusion model. Let $\vh_k$ after $k$ forward passes of the ADJUST module. Then they can be recursively defined as : 
\begin{equation}\label{eqn:unroll}
    \vh_{k+1} = g\left (\vh_{k} \, , \, \vx \oplus \tilde{x}^{\sigma(1)} \oplus \ldots \oplus \tilde{x}^{\sigma(k+1)} \right )
\end{equation}
where $\tilde{x}^{\sigma(k)}$ represents sampled token for position $\sigma(k)$. 
The algorithm for doing one denoising step is : (also refer Algo~\ref{alg:joint_sampling})\\
\textit{(1)} Do a forward pass of the base diffusion model and set $\vh_0 = f(\vx)$. \\
\textit{(2)} Sample the first token $\tilde{x}^{\sigma(1)}$ from base model's marginals $p^{\sigma(1)}(\cdot \mid \vx) = \mathrm{softmax}[\mW\vh^{\sigma(1)}_0]$ \\
\textit{(3)} Set the current unmasked string to $\vx \oplus \tilde{x}^{\sigma(1)}$ and current embedding to $\vh_0$. \\
\textit{(4)} Do forward pass of the ADJUST model $\vh_{1} \gets g\left (\vh_{0} \, , \, \vx \oplus \tilde{x}^{\sigma(1)} \right )$. \\
\textit{(5)} Sample the second token $\tilde{x}^{\sigma(2)}$ from new marginals $\mathrm{softmax}[\mW\vh^{\sigma(2)}_1]$. Repeat step 3-5 with updated counter. 

Note that the only difference between naive parallel and \method\ sampling is {updating the output embeddings} (line~\ref{diffline}). The final probability for sampling $K$ tokens in a single denoising step is:
\begin{align} \label{eqn:p_adjust}
& p_{\method}(\tilde{x}^{\sigma(1)}, \ldots, \tilde{x}^{\sigma(K)} \mid \vx) ~ = ~ \nonumber \\  &  \underbrace{p^{\sigma(1)} \left ( \tilde{x}^{\sigma(1)} \, | \, \vx  \right )}_{\mathrm{Sample\ token\ from\ base\ model\ probs}} \times \prod_{k=2}^{K} \, \underbrace{q^{\sigma(k)} \left ( \tilde{x}^{\sigma(k)} \mid \vh_{k-2} \, , \, \vx \oplus \tilde{x}^{\sigma(1)} \oplus \ldots \oplus \tilde{x}^{\sigma(k-1)} \right )}_{\mathrm{Sample\ tokens\ from\ ADJUST\ probs\ by\ iterating\ K-1\ times.}}    
\end{align}

\subsection{Training Methodology}\label{subsec:training}

\textbf{Training Loss.} 
We want our joint approximation $p_{\text{\method}}$ (Eqn~\ref{eqn:p_adjust}) to mimic the true joint distribution $p_*$ (Eqn~\ref{eqn:true_joint}). 
Notice that if $q \left ( \cdot \mid \vh_{k-2} \, , \, \vx \oplus \tilde{x}^{\sigma(1)} \oplus \ldots \oplus \tilde{x}^{\sigma(k-1)} \right )$ was equal to $p\left (\cdot \mid  \vx \oplus \tilde{x}^{\sigma(1)} \oplus \ldots \oplus \tilde{x}^{\sigma(k-1)}\right )$  for all values of $k\geq2$, $p_{\text{\method}}$ (Eqn~\ref{eqn:p_adjust}) would be exactly equal to $p_*$ (Eqn~\ref{eqn:true_joint}). 
Hence our training objective minimizes the KL divergence between $p$ and $q$. Our training loss is hence a sum of KL divergence terms between . 
\begin{equation}\label{eqn:training_loss}
    \mathcal{L}(\theta)~ = ~ \sum_{k=2}^K \,  \, \KL \left[p \left ( \cdot  \mid  \vx \oplus \ldots \oplus \tilde{x}^{\sigma(k-1)}\right ) \mid\mid q \left ( \cdot  \mid  \vh_{k-2}, \vx \oplus \ldots \oplus \tilde{x}^{\sigma(k-1)}\right ) \right]
\end{equation}
Note that probabilities $p(\cdot\mid\vx)$ and $q(\cdot\mid\vh,\vx)$ represent probabilities for each position in the sequence as $p^l$ and $q^l$. The KL divergence term is computed for each token position and not just for a single token position. The loss is represented in line ~\ref{algo_eqn_loss} of Algo~\ref{alg:dafter_training}.

\textbf{Generating Training Samples.} 
We describe how we sample partially masked token strings utilized in Eqn~\ref{eqn:training_loss} for training. 

\textbf{1)} We first realize that $q$ would be evaluated at test on the distribution of partially masked strings encountered during denoising with the base diffusion model $p$. We consider $\vx$ to be exactly the partially masked strings formed by the base model's denoising process. That is, given an initial distribution of masked strings, we first sample an initial \textit{prompt} string. We then run the base model's denoising process to completion. Note that here we are using the base model's unmasking logic (see App~\ref{app:unmasking_logic} for details) and perform single token $K=1$ unmasking to sample from the true joint. The history of all partially masked strings encountered during the denoising process serve as our distribution of $\vx$. 

\textbf{2)} Following the same logic, we argue that $\sigma$ for a training sample $\vx$ should be the next $K$ positions generated by the base model starting from $\vx$ (on single token unmasking). Finally we describe how to get $\vh_k$. Recall that, $\vh_k$ is defined as $g\left (\vh_{k-1} \, , \, \vx \oplus \tilde{x}^{\sigma(1)} \oplus \ldots \oplus \tilde{x}^{\sigma(k)} \right )$. We use the recursive computation of embeddings with teacher forcing of token inputs during training (line~\ref{algo_eqn_unrolling} of Algo~\ref{alg:dafter_training}). The ``unrolling" of $g$ during training is considered helpful by prior work ~\cite{liu2024deepseek, li2025eagle}.

Putting these things together, we present our training algorithm in Alg~\ref{alg:dafter_training}.



\section{Experiments}\label{sec:experiments}

\subsection{Experimental Setup}\label{subsec:experiment_setup}
We consider four different state-of-the-art discrete diffusion language models as base models for our empirical investigation. We consider a pretrained-only model (Dream-7B-Base, Llada-7B-Base) and an instruction-tuned model (Dream-7B-Instruct, Dream-7B-Coder)~\cite{ye2025dream}.
Recall that generating multiple tokens per denoising step leads to a deviation from the true underlying distribution. 
We therefore compare performances of different methods as a function of number of tokens $K$ generated for each denoising step. 
For simplicity of the setup, we generate the same number of tokens for each denoising step.
We use the base model's unmasking logic for our experiment (unless specified). Dream models' unmasking logic chooses positions with the least entropy and Llada uses the most confident token.  

\paragraph{Methods} We briefly describe the methods considered. 
\textbf{1)} \textit{True joint sampling}: This method corresponds to generating a single token $K=1$ in each denoising step. Recall that we established that generating one token per denoising step corresponds to sampling from the true joint distribution. Joint sampling serves as the oracle performance our method \method\ aims to mimic. 
\textbf{2)} \textit{Parallel sampling}: This is the naive parallel sampling, default in diffusion models. Each of the $K$ generated tokens is sampled independently from the base model's logits. 
\textbf{3)} \textit{Energy-based model sampling (EDLM)}~\cite{xu2024energy}: operates by drawing multiple token string unmasked completions (two in our experiments). It then chooses the completion with the lowest auto-regressive generative perplexity. The chosen string is then re-masked such that the final string has $K$ less masked positions (see Sec~\ref{app:hyperparameter} for discussion of hyper-parameter choices). We use the base model's unmasking logic. 
\textbf{4)} \textit{Dynamic Parallel Decoding (DPD)}~\cite{wu2025fast} : operates by unmasking all positions whose most likely token is above a confidence threshold. This method is suited for decoding in greedy (sampling temperature is 0) parallel settings. Comparison with DPD is presented in Sec~\ref{sec:runtime}
\textbf{5)} \textit{\method\ \textbf{(ours)}}: This is our joint approximation algorithm (Sec~\ref{sec:method}) which updates the logits conditioned on the sampled tokens. 
We acknowledge other baseline methods like \textit{adaptive parallel decoding}(APD)~\cite{israel2025accelerating} and \textit{discrete copula diffusion} (DCD)~\cite{liu2024discrete} which we omit from the main text.
Their formulation is orthogonal to our goal of mimic generation from the true joint distribution. 
(see Sec~\ref{app:baselines} for results and discussion).

\begin{table*}[t]
\centering
\begin{tabular}{lccccccccc}
\toprule
&  & \multicolumn{2}{c}{\textbf{K=1}} & \multicolumn{2}{c}{\textbf{K=2}} & \multicolumn{2}{c}{\textbf{K=4}} & \multicolumn{2}{c}{\textbf{K=8}} \\
& & NLL & Mauve & NLL & Mauve & NLL & Mauve & NLL & Mauve \\
\midrule
\multirow{3}{*}{\rotatebox[origin=c]{90}{T=0.6}} & Parallel & {1.11} & {1.00} & 1.61 & 0.89 & 2.21 & 0.29  & 2.21 & 0.02\\
 & EDLM & \textbf{0.83} & 0.98 & 1.33 & 0.94 &   1.89 &  0.31  & 2.05 & 0.03 \\
 & ADJUST \textbf{(ours)} & {1.11} & {1.00} & \textbf{1.23} & \textbf{0.97} & \textbf{1.64} & \textbf{0.79}  & \textbf{1.66} & \textbf{0.67}\\
\midrule
\multirow{3}{*}{\rotatebox[origin=c]{90}{T=1.0}} & Parallel & 4.25 & {1.00} & 5.68 & 0.76 & 6.52 & 0.31  & 7.10 & 0.19\\
 & EDLM & \textbf{3.75}  & 0.96 &  5.26 & 0.83 &  6.18 & 0.39 & 6.56 & 0.16 \\
& ADJUST \textbf{(ours)} & 4.25& {1.00} & \textbf{4.99} & \textbf{0.96} & \textbf{5.58} & \textbf{0.87} & \textbf{5.91} & \textbf{0.84} \\
\bottomrule
\end{tabular}
\caption{\label{table:nll_table} 
\textbf{Unconditional Generation. }In this table we compare the NLL and the MAUVE scores for three different methods : Marginal, Joint and Energy-based (EDLM). The base diffusion model is Dream-Base-7B.  
Across different number of tokens $K$ sampled per step, and different sampling temperature, joint sampling consistently outperforms both of the baselines, parallel sampling and EDLM. 
}
\end{table*}

\paragraph{Training Details} We train \method\ with a roll-out of three, that is, we assume $K=4$ for training. All models are trained for 2 epochs with constant learning rate of 5e-5 and batch size of 32. \method\ has 200M parameters (compared to 7B parameters for base model). 
Recall that \method\ assumes an initial distribution of prompts $\mathcal{D}$ (kindly refer Alg~\ref{alg:dafter_training}). 
For pretrained-only models we assume the initial prompt as just the beginning-of-sentence token, with max generation length of 128 tokens. We run the base model inference with a temperature of 1.0 for 100K samples to get our training dataset. For Dream-Instruct, we subsample 20K unique questions from MetamathQA~\cite{yu2023metamath} and generate with max generation length of 256 tokens, with temperature of 0.1 and nucleus of 0.9. For Dream-Coder we similarly subsample 20K prompts from Opencoder-LLM's opc-sft-stage2~\cite{Huang2024OpenCoderTO}. All data-generation runs take one GPU day on H100. All training runs take four GPU days on H100. Note that we are only sampling the prompts and not the ground truth responses. 

\subsection{Unconditional Generation}\label{subsec:language}

We consider generation with the input prompt as the beginning-of-sentence token, with a max generation length of 128. This \textit{unconditional generation} task aims to measure the language modeling capability of the model. 
We generate the output strings using our baseline methods and compare the negative log-likelihood (NLL) of the generated string as measured by an auto-regressive (AR) model (Qwen-2.5-7B for our setup). 
Recall that our target is to mimic the base model's true joint distribution. Hence we also measure the generated strings MAUVE score~\cite{mauve} (measured by GPT2-Large) with strings sampled from the true joint distribution, for given set of generation parameters. MAUVE score intuitively captures a notion of KL divergence between a two given sets of sentences and varies between zero and one with higher values implying a lower KL divergence between the distribution of the two sets. 

\begin{table*}[t]
\centering
\begin{tabular}{lcc|cccccc}
\toprule
\multirow{2}{*}{\textbf{Model}} & \multirow{2}{*}{\textbf{Task}} & \multirow{2}{*}{\textbf{K=1}} & \multicolumn{2}{c}{\textbf{K=2}} & \multicolumn{2}{c}{\textbf{K=4}} & \multicolumn{2}{c}{\textbf{K=8}}  \\
& &  & Parallel & ADJ. & Parallel & ADJ. & Parallel & ADJ. \\
\midrule
\multirow{3}{*}{\rotatebox[origin=c]{0}{Llada-Base}}  & GSM & 68.94 & 60.61 & 66.67 & 54.55 & 57.58 & 19.70 & 28.79 \\
& MBPP & 40.40 & 28.40 & 37.40 & 19.40 & 29.40 & 6.60 & 20.00 \\
& HEval & 33.54 & 25.00 & 29.88 & 12.80 & 23.78 & 5.49 & 10.37 \\
\midrule
\multirow{2}{*}{\rotatebox[origin=c]{0}{Dream-Base}} & GSM & 78.40 & 71.20 & 75.60 & 43.20 & 64.00 & 11.60 & 36.80 \\
& MBPP & 48.00 & 40.80 & 44.80 & 33.60 & 33.60 & 18.40 & 19.60 \\
\midrule
\multirow{2}{*}{\rotatebox[origin=c]{0}{Dream-Inst}} & GSM & 85.20 & 76.80 & 80.08 & 58.00 & 74.00 & 20.00 & 44.40 \\
& MATH & 32.57 & 32.29 & 31.42 & 15.14 & 22.28 & 1.14 & 5.70 \\
\midrule
\multirow{2}{*}{\rotatebox[origin=c]{0}{Dream-Coder}} & MBPP & 63.00 & 47.00 & 57.40& 12.60 & 16.00 & 4.60 & 9.20 \\
& HEval & 75.00 & 55.49 & 73.17 & 20.73 & 57.32 & 1.83 & 28.05 \\
\bottomrule
\end{tabular}
\caption{\label{table:combined_table}
\textbf{Downstream Evaluation.} In this table we report numbers on downstream tasks, GSM8k \& MBPP for the Dream-Base (denoted as Base). Since \method\ for Dream-Instruct (denoted as Inst.) in trained on MetamathQA prompts, we report GSM8k \& MATH numbers. Sampling a single token for each denoising step gives the best performance. \method\ achieves better downstream accuracy tasks across values of tokens sampled in each denoising step $K$. 
}
\vspace{-2em}
\end{table*}

\paragraph{Results} are presented in Table~\ref{table:nll_table}. 
We present results for pre-trained only Dream-Base-7B. We evaluate generations for two sampling temperatures, 0.6 and 1.0, highlighting the robustness of our method to sampling parameters. The MAUVE score is computed w.r.t. the samples from diffusion model's true joint distribution, that is, samples generated with one token per diffusion step $K=1$ sampling. 
For $K=1$, EDLM has lower NLL compared to our method, while (seemingly counterintuitively) having lower MAUVE. This is an expected behavior, as EDLM samples two different completions for each denoising step and chooses the completion with a lower NLL. 
The selection of lower NLL completion biases the EDLM model away from the true joint distribution of the diffusion model and more towards the underlying distribution of the auto-regressive model (used for computing NLL). 
This is also why EDLM achieves lower NLL than parallel and \method\ for $K=1$. 
Results for $K=8$ show how well \method\ generalizes beyond the training rollouts. 

\subsection{Downstream Evaluation}\label{subsec:downstream}
Here we show that our \method\ method leads to better downstream accuracy than naive parallel sampling and closes matches performance of $K=1$ for any degree of parallel generation.  
We use low temperature sampling (0.0-0.1) for generation. 
Due to computational constraints we limit our evaluation to 250 question from GSM8K \& MBPP and 350 question (50 question per task) for Minerva math. 
Exact sampling and evaluation parameters are summarized in Sec~\ref{app:sampling_params}. 

\paragraph{Results} Recall that the \method\ for the pretrained-only model is trained on domain-agnostic data (generations starting from begin-of-sentence token). 
Hence the results for Dream-Base and Llada-Base show the utility of a generically trained \method\ for downstream tasks like GSM8K and MBPP in Table~\ref{table:combined_table}. Our method is better than marginal across all values of number of generated tokens in parallel even without ever specifically being prompted for math or code data.  
Since the downstream evaluation utilizes low temperature sampling, EDLM leads to same performance as naive sampling. Hence we don't report those numbers in the table.  
\method\ for Dream-Instruct model is trained on MetamathQA prompts. Hence we evaluate on math domain downstream tasks (GSM8K and Minerva-math). Similarly, since we train Dream-Coder on opc-sft-stage2 prompts, we evaluate it on coding tasks. Results for $K=8$ lie outside the training rollout range of $K=4$ and demonstrate the generalization of \method\ along longer prediction horizons. 

\subsection{Throughput evaluation}\label{sec:runtime}

We report numbers on the accuracy of various methods as a function of throughout for Dream-Base-7B model. Evaluations are done on a single H100 GPU with batch size one. For parallel and ADJUST methods, we vary the number of tokens sampled per diffusion denoising step to control the token throughput. Tokens sampled per denoising step varies from one to six. For DPD, we vary the confidence threshold for accepting tokens from 1.0 to 0.4. Recall that DPD is a greedy (temperature=0) sampling based method and hence we omit unconditional generation results for DPD, which requires sampling. 

\textbf{Results} are encapsulated in the caption of Fig~\ref{fig:runtime}.
We would like to emphasize that ADJUST for Dream-Base is trained in a downstream agnostic manner. 
The discrepancy between tokens/sec between the unconditional generation and GSM8k plots can be attributed to the longer eight-shot prompt used for GSM8k evaluation. 


\begin{figure*}[t]
    \centering
    \includegraphics[width=\linewidth]{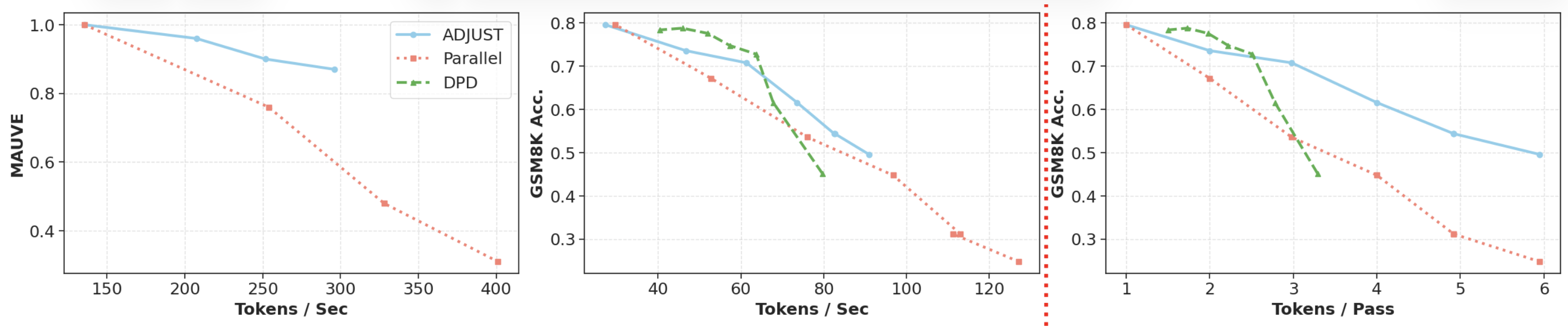}
    \caption{\label{fig:runtime} 
    \textbf{Throughput evaluation. }In this figure we report tokens produced per second for unconditional (leftmost) and downstream evaluation (middle) on GSM8k for Dream-Base-7B (Sec~\ref{sec:runtime}). 
    Recall that DPD is a greedy (temperature=0) sampling based method and hence is missing from unconditional generation figure which requires sampling to generate text from base model's underlying distribution.
    For both evaluation settings, ADJUST demonstrates gain in performance for a given throughput compared to naive parallel sampling. Additionally, while DPD is limited to a smaller range of tokens produced per second, ADJUST achieves higher throughput (>80 toks/sec) while outperforming both parallel sampling and DPD in that regime. The rightmost figure shows comparison with numbers with tokens per pass on the x-axis
    }
    \vspace{-1em}
\end{figure*}
\begin{wrapstuff}[r, width=0.60\linewidth, type=figure]
    \centering
    \begin{table}[H]
    \centering
    \begin{tabular}{lccccccc}
    \toprule
     & W2S(E) & W2S(M) & W2S(H)  \\
    \midrule
    True Joint & \underline{82.0} & \textbf{87.0} & \textbf{69.0} \\
    Parallel (K=4) & 81.0 & 77.0 & 63.0\\
    ADJUST (K=4) & \textbf{84.0} & \underline{85.0} & \textbf{69.0} \\
    \bottomrule
    \end{tabular}
    \caption{\label{tab:main_parallelbench} In this table, we report numbers on the word-to-sentence (W2S) task of ParallelBench~\cite{kang2025parallelbench}.
    }
    \end{table}
\end{wrapstuff}

\subsection{ParallelBench}\label{subsec:ar_decoding}

ParallelBench~\cite{kang2025parallelbench} is a recent benchmark that quantifies the quality degradation under parallel decoding with diffusion LMs. 
We here report numbers for a representative task word-to-sentence using Dream-Base-7B as the base diffusion model in Table~\ref{tab:main_parallelbench}. The task prompt asks the model to generate a coherent sentence with a set of given words present in the sentence. A generation is valid if it's both coherent (grammatically) and includes all the specified words. We report the fraction of valid generation for three task difficulty levels (easy (E), medium (M) and hard (H)). We kindly refer the reader to App. Table~\ref{tab:parallelbench}) for results on the other tasks in parallelbench.

\section{Conclusion, Limitations and Future Work}
\label{sec:conclusion}
\textbf{Conclusion.} We study the problem of jointly sampling multiple tokens in each diffusion denoising step for masked diffusion LMs. Our main contribution was the formalization of our approximate joint sampler \method, along with proposing a new training methodology based on sampling trajectory. For \textbf{future work}, we hope to design systems which can help \method\ realize it's gains on token per pass to similar gains on tokens per second. 
\textbf{Limitations.} Since the only open-source diffusion LLM exist at 7B scale, we don't have ablations on base model size. For the same reasons, we only consider masked diffusion model as all open-source diffusion build on this paradigm. 

\bibliographystyle{plain}
\bibliography{example_paper}


\appendix

\section{Additional Discussions}

\subsection{Hyper-parameter selection}\label{app:hyperparameter}
\paragraph{EDLM} We use Qwen-2.5-0.5B to evaluate generative perplexity of sampled strings and sample two string completions. For EDLM we choose just two samples as this was shown to be optimal number presented in the EDLM paper~\cite{xu2024energy}. We ablated with 4 choices as well but got similar results. Hence we just consider sampling 2 options. 
Another consideration for EDLM is the choice of the auto-regressive model. We take the model to be Qwen2.5-0.5B. We also tried Qwen2.5-7B but that also didn't help improve the performance much. The numbers are reported as EDLM (Big) in Table~\ref{table:EDLM_table}.

\section{Additional Baselines}\label{app:baselines}

\subsection{Discrete Copula Diffusion}
DCD uses an auto-regressive model of the same size and distribution as the base diffusion model to generate a complete unmasked token string. An auto-regressive generation strategy uses $|\mathcal{M}(\vx)|$ forward passes of the AR model (of the same size as the diffusion model). 
We can alternatively use $|\mathcal{M}(\vx)|$ forward passes of the diffusion model (with $K=1$) to directly sample from the true joint (without using the AR model). 
DCD~\citet{liu2024discrete} aims to construct a joint distribution over the whole token string instead of just tokens to be unmasked. Since they have a joint over all the masked positions, their method works in a regime of aggressive unmasking.  
\method\ does auto-regressive generation over only the tokens to be unmasked. Hence, our method (intuitively) predicts a joint over positions to be unmasked and not the whole sentence. This makes our method much more computationally friendly then DCD.

\subsection{Adaptive parallel decoding} (APD) ~\cite{israel2025accelerating}: is a specialized generation technique restricted to left-to-right decoding. At each denoising step APD samples a complete unmasked token string (in parallel). It then uses an auto-regressive evaluator (model) to adaptively selects a prefix string to commit and unmask. Because APD generates variable number of tokens in each denoising step and is strictly left-to-right.

We first note that APD is a specialized left-to-right constrained decoding algorithm suited for reasoning heavy tasks evaluated at low temperatures. 
For e.g., for Dream-7B-Instruct, using APD we can generate (on an average) upto five tokens per denoising iteration for low temperature (=0.1) sampling on GSM8k prompts with negligible loss in performance.
Here we investigate the performance of methods for open-ended tasks like unconditional generation and high temperature GSM8k sampling. 
We acknowledge the synthetic nature of high temperature GSM8k sampling and emphasize that these numbers are for completeness. 


\textbf{Results} are encapsulated in caption of Fig~\ref{fig:apd}. We highlight that we use left-to-right decoding for APD and use base model's unmasking logic for parallel and \method. The results show that \method\ is applicable across tasks and sampling temperatures where APD fails. 
\begin{figure*}[h]
    \centering
    \begin{subfigure}{0.49\linewidth}
        \centering
        \includegraphics[width=\linewidth]{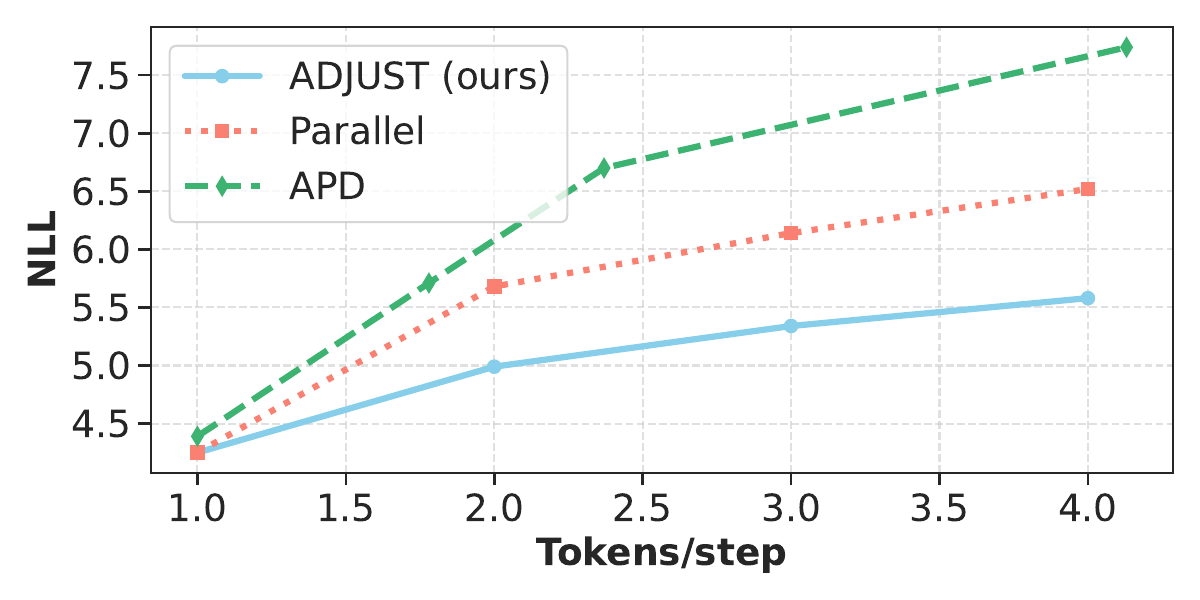}
    \end{subfigure}
    \hfill
    \begin{subfigure}{0.49\linewidth}
        \centering
        \includegraphics[width=\linewidth]{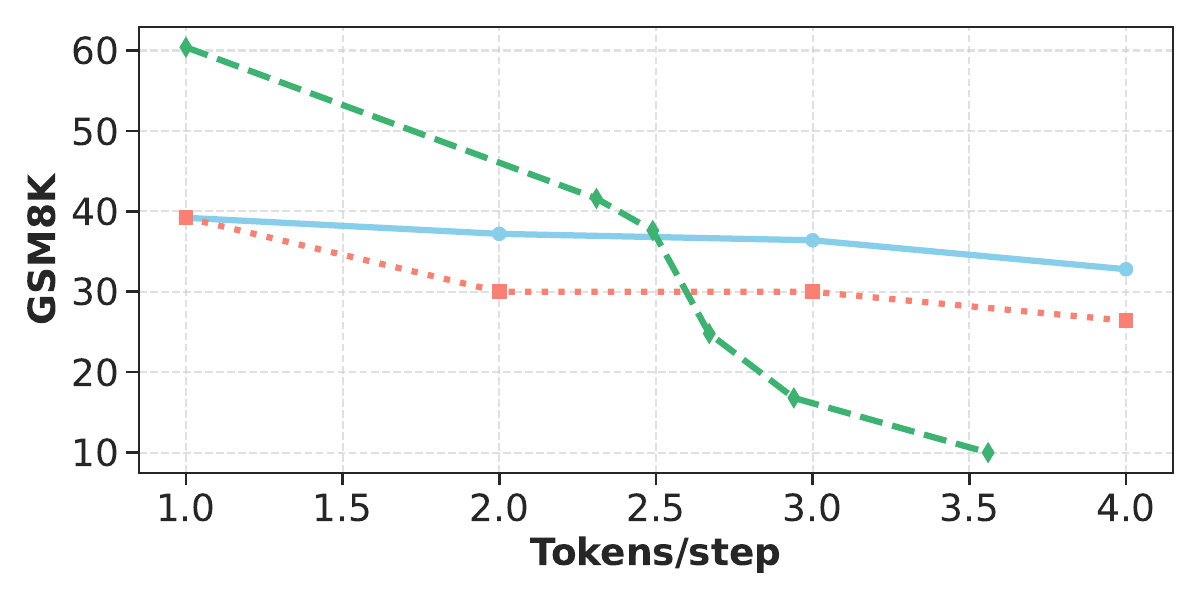}
    \end{subfigure}
    \caption{\label{fig:apd} In this figure we compare adaptive parallel decoding (working left-to-right) to naive parallel decoding and \method.
    Sampling $K>1$ tokens per diffusion step, for left-to-right decoding shows a steeper increase in NLL (left) than using the base model's unmasking logic.   
    Hence APD generation is not suited for such open-ended tasks and under-performs. 
    The right figure shows GSM8k evaluation for high temperature sampling (temperature=1.0) for Dream-Instruct model. Left-to-right generation is naturally good for math reasoning tasks, as shown by higher accuracy under APD when compared to naive/joint for $K=1$ tokens per step. APD leads to a quicker decrease in accuracy when compared to baseline and our method. 
    }
\vspace{-0.5cm}
\end{figure*}

\section{Additional Experiments from omitted from the main text}
\subsection{Complete NLL and MAUVE numbers}
See Table~\ref{table:all_nll_table} for all numbers

\begin{table}[H]
\centering
\begin{tabular}{lccccccccc}
\toprule
&  & \multicolumn{2}{c}{\textbf{K=1}} & \multicolumn{2}{c}{\textbf{K=2}} & \multicolumn{2}{c}{\textbf{K=3}} & \multicolumn{2}{c}{\textbf{K=4}} \\
& & NLL & Mauve & NLL & Mauve & NLL & Mauve & NLL & Mauve \\
\midrule
\multirow{3}{*}{\rotatebox[origin=c]{90}{T=0.6}} & Parallel & {1.11} & {1.00} & 1.61 & 0.89 & 2.18 & {0.57} & 2.21 & 0.29 \\
 & EDLM & \textbf{0.83} & 0.98 & 1.33 & 0.94 &  1.85 & 0.54 &  1.89 &  0.31 \\
 & ADJUST \textbf{(ours)} & {1.11} & {1.00} & \textbf{1.23} & \textbf{0.97} & \textbf{1.54} & \textbf{0.82} & \textbf{1.64} & \textbf{0.79} \\
\midrule
\multirow{3}{*}{\rotatebox[origin=c]{90}{T=1.0}} & Parallel & 4.25 & {1.00} & 5.68 & 0.76 & 6.14 & 0.48 & 6.52 & 0.31 \\
 & EDLM & \textbf{3.75}  & 0.96 &  5.26 & 0.83 &  5.76 & 0.58 & 6.18 & 0.39 \\
& ADJUST \textbf{(ours)} & 4.25& {1.00} & \textbf{4.99} & \textbf{0.96} & \textbf{5.34} &\textbf{ 0.90} & \textbf{5.58} & \textbf{0.87} \\
\midrule
&  & \multicolumn{2}{c}{\textbf{K=5}} & \multicolumn{2}{c}{\textbf{K=6}} & \multicolumn{2}{c}{\textbf{K=7}} & \multicolumn{2}{c}{\textbf{K=8}} \\
& & NLL & Mauve & NLL & Mauve & NLL & Mauve & NLL & Mauve \\
\midrule
\multirow{3}{*}{\rotatebox[origin=c]{90}{T=0.6}} & Parallel & 2.00 & 0.12 & 2.11 & 0.07 & 2.19 & 0.05 & 2.21 & 0.02\\
 & EDLM & 1.87 & 0.14 & 1.95 & 0.07 & 2.03 & 0.05 & 2.05 & 0.03 \\
 & ADJUST \textbf{(ours)} & \textbf{1.47} &\textbf{ 0.79} & \textbf{1.54} &\textbf{ 0.77} & \textbf{1.60} & \textbf{0.76} & \textbf{1.66} & \textbf{0.67}\\
\midrule
\multirow{3}{*}{\rotatebox[origin=c]{90}{T=1.0}} & Parallel & 6.78 & 0.22 & 6.80 & 0.24 & 6.89 & 0.17 & 7.10 & 0.19\\
 & EDLM & 6.19 & 0.27 & 6.26 & 0.26 & 6.35 & 0.21 & 6.56 & 0.16 \\
& ADJUST \textbf{(ours)} &\textbf{ 5.67} & \textbf{0.87} & \textbf{5.76} &\textbf{ 0.85} & \textbf{5.83} & \textbf{0.85} & \textbf{5.91} & \textbf{0.84} \\
\bottomrule
\end{tabular}
\caption{\label{table:all_nll_table} In this table we compare the NLL and the MAUVE scores for three different methods : Marginal, Joint and Energy-based (EDLM). 
Across different number of tokens $K$ sampled per step, and different sampling temperature, joint sampling consistently outperforms both of the baselines, parallel sampling and EDLM. 
}
\end{table}

\subsection{Complete ParallelBench Numbbers}
See Table~\ref{tab:parallelbench}
\begin{table}[H]
\centering
\begin{tabular}{lccccccc}
\toprule
Model & W2S(E) & W2S(M) & W2S(H) & Copy & Ins(idx) & Ins(rand) & Rm(idx) \\
\midrule
Dream (x1) & 82.0 & 87.0 & 69.0 & 73.0 & 30.0 & 42.0 & 72.0 \\
Parallel (x4) & 81.0 & 77.0 & 63.0 & 55.0 & 12.0 & 13.0 & 57.0 \\
ADJUST (x4) & 84.0 & 85.0 & 69.0 & 63.0 & 21.0 & 20.0 & 62.0 \\
\midrule
Model & Rm(rand) & Rpl(idx) & Rpl(rand) & Reverse & Shuffle & Sort & \\
\midrule
Dream (x1) & 62.0 & 51.0 & 52.0 & 68.0 & 62.0 & 21.0 & \\
Parallel (x4) & 36.0 & 35.0 & 1.0 & 27.0 & 24.0 & 13.0 & \\
ADJUST (x4) & 39.0 & 46.0 & 3.0 & 39.0 & 14.0 & 16.0 & \\
\bottomrule
\end{tabular}
\caption{\label{tab:parallelbench}Evaluation on ParallelBench using Dream-Base-7B model}
\end{table}

\subsection{EDLM Experiments}

\begin{table}[H]
\centering
\begin{tabular}{l|ccc|ccc}
\toprule
 & \multicolumn{3}{c|}{\textbf{T=0.6}}   & \multicolumn{3}{c}{\textbf{T=1.0}}  \\
& Spec=1 & Spec=2 & Spec=3 & Spec=1 & Spec=2 & Spec=3 \\
\midrule
Tok/sec &  35.57 & 52.87 & 71.00 & 35.56 & 52.91 &  70.97\\
\bottomrule
\end{tabular}

\vspace{0.5cm}

\begin{tabular}{l|ccc|cccc}
\toprule
 & \multicolumn{3}{c|}{\textbf{T=0.6}}   & \multicolumn{3}{c}{\textbf{T=1.0}}  \\
& Spec=1 & Spec=2 & Spec=3 & Spec=1 & Spec=2 & Spec=3 \\
\midrule
EDLM (Small) & 1.33 &  1.85&  1.89 & 5.26 &  5.76& 6.18 \\
EDLM (Big)  & 1.61 & 2.08  & 2.08 & 5.27 & 5.74 & 6.18  \\
\bottomrule
\end{tabular}

\caption{\label{table:EDLM_table} We report the runtime of the EDLM algorithm (with Qwen-2.5-0.5B) for experimental setup in Table~\ref{table:nll_table} in Tokens/sec. These numbers are comparable to those in figure 1. In this table we report the negative log-likelihood numbers on decoding with EDLM.  
We fix the generation length to be 128 tokens and nucleus sampling to be 0.95, and temperature to be 1.0.
Here we use Qwen2.5-7B as the verifier for EDLM (Big), i.e., the same model which is used to evaluate responses, while using Qwen2.5-0.5B as the verifier for EDLM (Small). We can see identical results because at each position we sample only two options both of which are identically rated by both the small and the big models  
}
\end{table}

\subsection{Architecture}
See Fig~\ref{fig:joint_architecture}.

\begin{figure}[h]  
    \centering
    \includegraphics[width=0.3\textwidth]{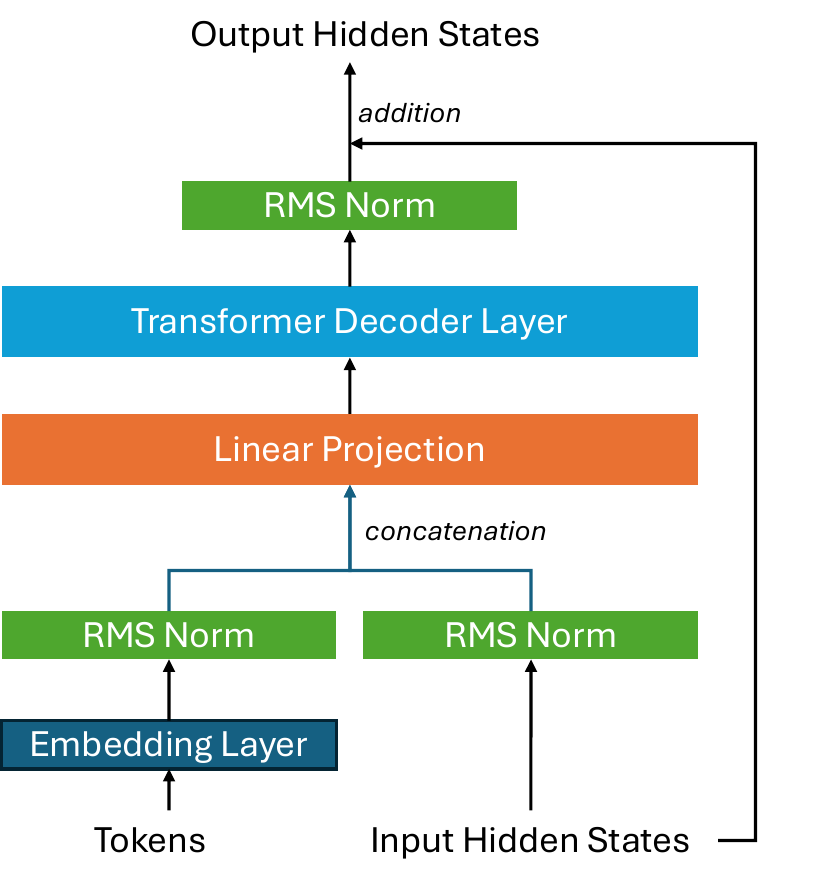}
    \caption{This figure shows the architecture used for our draft model}
    \label{fig:joint_architecture}
\end{figure}

\section{Additional Experimental Details}

\subsection{Sampling Parameters}\label{app:sampling_params}

For downstream evaluation, we take parameters from the official Dream repository\footnote{\url{https://github.com/DreamLM/Dream}}

\paragraph{Pretrained-only (Dream-Base/Llada-Base)} For NLL and MAUVE evaluation we use temperature in \{1.0, 0.6\} and topp of 1.0 with generation length of 128. We summarize them here for completeness. For GSM8K evaluation we use temperature of 0.0 and 256 generation length with 8 shot examples in context. For MBPP we use 0.2 temperature, topp of 1.0 with generation length of 512 and num few shot of 3 examples. 

\paragraph{Dream-7B-Instruct} For GSM8K we use 0.1 temperature, 0.9 topp, and generation length of 256. Also we use the chat template for evaluation. For Minerva-math we use 4 few shot, 512 generation length and 0.1 temperature and 0.9 topp. Our evaluation code for Minerva relies on an older version from lm-evaluation-harness that doesn't utilize ``math\_verify" and hence the numbers might be under-reported here. 

\paragraph{Dream-7B-Coder} For HumanEval we use 0.2 temperature, 0.9 topp, and generation length of 512 and four few-shot examples. Similarly for MBPP we use the same sampling parameters

\section{Unmasking Logic}\label{app:unmasking_logic}
In our discussion till now, we assumed the list of positions to be generated $\sigma$ to be given to us. For the current state-of-the-art diffusion LLMs, positions to be generated next are decided by an operator which is a function of the logits, denoted as TOP. Base diffusion language models specify their criterion to decide the most suitable position to generate next. For example, Dream models~\cite{ye2025dream} select the position with the least entropy to generate next, Llada~\cite{nie2025large} unmasks position with the highest confidence. \method\ is agnostic to the choice of the unmasking logic and uses the base model's unmasking logic for generation. A subtle difference to note is that naive parallel sampling of $K$ tokens calls TOP on the same output embeddings $f(\vx)$, $K$ times, yielding top $K$ positions with (say) least entropy in $\mW f$. \method\ uses the updated logits to get the index of the next position, $\sigma(k) = \text{TOP} \left \{ \mW \vh_{k} \right\}$, which we qualitatively observe are often different than using the base model's logits $\mW f$. 


\newpage
\section*{NeurIPS Paper Checklist}



\begin{enumerate}

\item {\bf Claims}
    \item[] Question: Do the main claims made in the abstract and introduction accurately reflect the paper's contributions and scope?
    \item[] Answer: \answerYes{} 
    \item[] Justification: All claims in the abstract are accurately represented in the paper

\item {\bf Limitations}
    \item[] Question: Does the paper discuss the limitations of the work performed by the authors?
    \item[] Answer: \answerYes{} 
    \item[] Justification: We have Sec~\ref{sec:conclusion} in the paper which discusses our limitations

\item {\bf Theory assumptions and proofs}
    \item[] Question: For each theoretical result, does the paper provide the full set of assumptions and a complete (and correct) proof?
    \item[] Answer: \answerNA{} 
    \item[] Justification: We don't have any theoretical results. 

    \item {\bf Experimental result reproducibility}
    \item[] Question: Does the paper fully disclose all the information needed to reproduce the main experimental results of the paper to the extent that it affects the main claims and/or conclusions of the paper (regardless of whether the code and data are provided or not)?
    \item[] Answer: \answerYes{} 
    \item[] Justification: Yes, all details are shared. Work is reproducible. 

\item {\bf Open access to data and code}
    \item[] Question: Does the paper provide open access to the data and code, with sufficient instructions to faithfully reproduce the main experimental results, as described in supplemental material?
    \item[] Answer: \answerNo{} 
    \item[] Justification: The dataset is open-source. We use open-source repository for experimentation and can be done easily. 

\item {\bf Experimental setting/details}
    \item[] Question: Does the paper specify all the training and test details (e.g., data splits, hyperparameters, how they were chosen, type of optimizer) necessary to understand the results?
    \item[] Answer: \answerYes{} 
    \item[] Justification: Experimentation details are there in Sec~\ref{sec:experiments}.

\item {\bf Experiment statistical significance}
    \item[] Question: Does the paper report error bars suitably and correctly defined or other appropriate information about the statistical significance of the experiments?
    \item[] Answer: \answerNo{} 
    \item[] Justification: The standard errors are small for all the numbers reported. For ease of exposition we have not included them. 

\item {\bf Experiments compute resources}
    \item[] Question: For each experiment, does the paper provide sufficient information on the computer resources (type of compute workers, memory, time of execution) needed to reproduce the experiments?
    \item[] Answer: \answerYes{} 
    \item[] Justification: Yes, compute requirements are in the main paper Sec~\ref{sec:experiments}.
    
\item {\bf Code of ethics}
    \item[] Question: Does the research conducted in the paper conform, in every respect, with the NeurIPS Code of Ethics \url{https://neurips.cc/public/EthicsGuidelines}?
    \item[] Answer: \answerYes{} 
    \item[] Justification: We conform to the code of ethics. 

\item {\bf Broader impacts}
    \item[] Question: Does the paper discuss both potential positive societal impacts and negative societal impacts of the work performed?
    \item[] Answer: \answerNA{} 
    \item[] Justification: Our work advances language models and there are no additional malicious use-cases to the best of our knowledge, beyond using better language models for harm. 
    
\item {\bf Safeguards}
    \item[] Question: Does the paper describe safeguards that have been put in place for responsible release of data or models that have a high risk for misuse (e.g., pre-trained language models, image generators, or scraped datasets)?
    \item[] Answer: \answerNo{} 
    \item[] Justification: No such risks

\item {\bf Licenses for existing assets}
    \item[] Question: Are the creators or original owners of assets (e.g., code, data, models), used in the paper, properly credited and are the license and terms of use explicitly mentioned and properly respected?
    \item[] Answer: \answerNA{} 
    \item[] Justification: All the data used is open-source and appropriate citations are given. 

\item {\bf New assets}
    \item[] Question: Are new assets introduced in the paper well documented and is the documentation provided alongside the assets?
    \item[] Answer: \answerNA{} 
    \item[] Justification: No new assets 

\item {\bf Crowdsourcing and research with human subjects}
    \item[] Question: For crowdsourcing experiments and research with human subjects, does the paper include the full text of instructions given to participants and screenshots, if applicable, as well as details about compensation (if any)? 
    \item[] Answer: \answerNA{} 
    \item[] Justification: No crowdsourced research

\item {\bf Institutional review board (IRB) approvals or equivalent for research with human subjects}
    \item[] Question: Does the paper describe potential risks incurred by study participants, whether such risks were disclosed to the subjects, and whether Institutional Review Board (IRB) approvals (or an equivalent approval/review based on the requirements of your country or institution) were obtained?
    \item[] Answer: \answerNA{} 
    \item[] Justification: No such research 

\item {\bf Declaration of LLM usage}
    \item[] Question: Does the paper describe the usage of LLMs if it is an important, original, or non-standard component of the core methods in this research? Note that if the LLM is used only for writing, editing, or formatting purposes and does \emph{not} impact the core methodology, scientific rigor, or originality of the research, declaration is not required.
    \item[] Answer: \answerNA{} 
    \item[] Justification: LLMs are not used. 

\end{enumerate}

\end{document}